\documentclass[conference]{IEEEtran}
\IEEEoverridecommandlockouts
\usepackage{cite}
\usepackage{amsmath,amssymb,amsfonts}
\usepackage{algorithmic}
\usepackage{graphicx}
\usepackage{textcomp}
\usepackage[table, dvipsnames]{xcolor}
\usepackage{authblk}
\usepackage{subcaption}
\usepackage{tikz}
\usepackage{multirow}
\usepackage[colorlinks=true, 
            linkcolor=blue, 
            citecolor=red, 
            urlcolor=blue]{hyperref}
\IEEEoverridecommandlockouts

\def\BibTeX{{\rm B\kern-.05em{\sc i\kern-.025em b}\kern-.08em
    T\kern-.1667em\lower.7ex\hbox{E}\kern-.125emX}}
\begin{document}

\makeatletter
\def\ps@IEEEtitlepagestyle{%
  \def\@oddhead{%
    \parbox{\textwidth}{\centering
      \vspace{1em}
      \large\textcolor{gray}{This paper has been accepted to IEEE International Conference on Multimedia \& Expo 2025}
      \vspace{1em}
    }%
  }%
  \def\@oddfoot{}%
}
\makeatother

\title{Towards End-to-End Neuromorphic Event-based 3D Object Reconstruction Without Physical Priors\vspace{-0.2em}}



\author{
 \textbf{\textit{Chuanzhi Xu$^{\dagger}$, Langyi Chen, Haodong Chen, Vera Chung, Qiang Qu}} \\
 School of Computer Science, The University of Sydney, NSW, Australia\vspace{-1.em}
}
\maketitle
\renewcommand{\thefootnote}{}
\footnotetext{$^{\dagger}$ Corresponding author. Email: \href{mailto:chxu4146@uni.sydney.edu.au}{chxu4146@uni.sydney.edu.au}}

\begin{abstract}
Neuromorphic cameras, also known as event cameras, are asynchronous brightness-change sensors that can capture extremely fast motion without suffering from motion blur, making them particularly promising for 3D reconstruction in extreme environments. However, existing research on 3D reconstruction using monocular neuromorphic cameras is limited, and most of the methods rely on estimating physical priors and employ complex multi-step pipelines. In this work, we propose an end-to-end method for dense voxel 3D reconstruction using neuromorphic cameras that eliminates the need to estimate physical priors. Our method incorporates a novel event representation to enhance edge features, enabling the proposed feature-enhancement model to learn more effectively. Additionally, we introduced \textit{Optimal Binarization Threshold Selection Principle} as a guideline for future related work, using the optimal reconstruction results achieved with threshold optimization as the benchmark. Our method achieves a 54.6\% improvement in reconstruction accuracy compared to the baseline method.
\end{abstract}


\section{Introduction}
\label{sec:intro}
3D reconstruction in VR/AR applications enables realistic restoration of scenes and objects, serving as a 3D form of information present that provides users with a more immersive experience. Many devices can be used to collect data for 3D reconstruction, including traditional RGB cameras, RGB-D cameras, LiDAR, structured light systems, etc. However, they have different limitations, including limited dynamic range, motion blur, high power consumption, etc.~\cite{[T7]}. 

Neuromorphic cameras, also known as event cameras, are bio-inspired sensors responding to local brightness changes~\cite{[T7]}. Each pixel in a neuromorphic camera operates independently and asynchronously, which differs from traditional (frame-based) RGB cameras that capture pictures with a shutter and have a fixed interval in recording video~\cite{[T7], [T13-12], qu2024e2hqv}. Neuromorphic cameras report changes in brightness only when a threshold is reached. When there is a greater change in brightness in the scene or object, such as when the object is moving faster, more event data will be generated. The neuromorphic camera produces a continuous stream of events, which includes the coordinates, precise timestamp, and the polarity of brightness change.
\begin{figure*}[t]
    \centering
    \includegraphics[width=\linewidth]{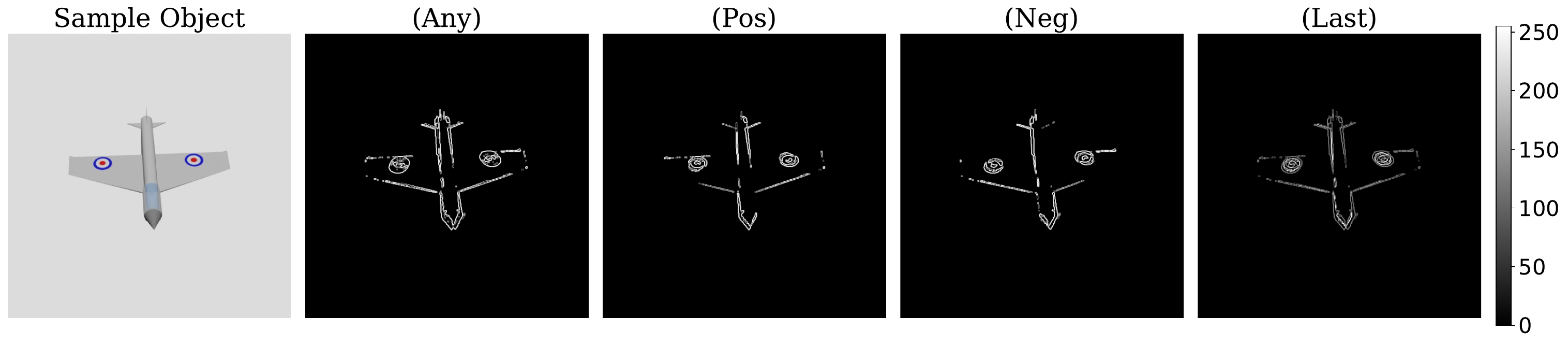} 
    \caption{Visualizations of different modes of \textit{Sobel Event Frame} applied on event stream of an airplane object.}
    \label{fig:sobel}
\end{figure*}

Neuromorphic cameras in 3D reconstruction tasks can be divided into stereo and monocular types \cite{[survey]}. Stereo neuromorphic cameras involve multiple rigidly connected neuromorphic cameras, providing information from multiple viewpoints. These methods typically perform scene scanning and produce real-time semi-dense reconstruction results. They often follow the classical two-step stereo solution: matching disparities at the same timestamps and then triangulating 3D points to calculate the depth information of every point in the scene \cite{[T5],[T7]}. However, performing 3D reconstruction with a monocular neuromorphic camera, which lacks disparity information, often requires more complex computations. Such methods must be combined with the physical prior knowledge (e.g., camera trajectories) to achieve similar results as parallax in the stereo task. The physical prior is always required, and it can either be obtained by simultaneous mapping with other devices or be predicted by a Visual Odometer (VO) or SLAM algorithm targeting the event stream \cite{[T5]}. Moreover, these methods all require a complex event-to-3D pipeline \cite{[T10], [T13], [T14], [T9]}. The pipeline is a full workflow that optimally processes event data and reconstructs it into a 3D model, including steps such as event representation, prior estimation, computation of parallax, triangulation, depth estimation, and 3D model reconstruction \cite{[survey]}. Event processing pipelines are a common research direction in these tasks. A recent study introduced a neural network, E2V \cite{[Dense]}, capable of directly taking the represented event stream as input and outputting voxel results. This approach made progress in simplifying the event processing pipeline. It is also the first method to perform voxel-based 3D reconstruction using a monocular neuromorphic camera. However, it leaves substantial room for improvement in reconstruction accuracy.

In this research, we propose an end-to-end method for dense voxel 3D reconstruction using neuromorphic cameras that eliminates the need to estimate physical priors. Our contributions can be summarized as:
\begin{itemize}
    \item We propose a novel event representation method, \textit{Sobel Event Frame}, which enhances edge features and restrains redundant data in the event stream, enabling effective learning of 3D features.
    \item We propose a 3D reconstruction model designed to enhance the learning of edge features in event streams with Efficient Channel Attention, building on the approach of event-based 3D reconstruction without the need for physical priors or pipelines.
    \item We propose \textit{Optimal Binarization Threshold Selection Principle} and suggest it as a guideline for future research about event-based 3D reconstruction with deep learning.
\end{itemize}

\section{Related Work}
\subsection{Physics and Geometry-based Methods}
Most methods for performing spatial scanning and instantaneous 3D reconstruction using a monocular neuromorphic camera are based on physical and geometric computations. These methods establish strict event processing pipelines, requiring steps such as feature extraction and feature matching, and must compute physical prior information. 

Kim et al. proposed the first method for depth estimation using a moving monocular neuromorphic camera \cite{[T6]}, which mainly consists of three decoupled probabilistic filters that estimate the 6-DoF motion of the camera, the scene intensity gradient, and the inverse depth of the scene relative to keyframes. Rebecq et al., in 2016, proposed an event-based visual odometry algorithm (EVO) \cite{[T9]}. The tracking module in EVO uses Event Frame to represent event streams, generate edge images, and estimate the camera pose. Its mapping module expands the semi-dense 3D map when new events arrive and feeds the map back into the tracking module. The EMVS method proposed by Rebecq et al. in 2018, based on a known camera trajectory \cite{[T5]}, back-projects events into space to create a light-density volume \cite{[T61]}. It identifies the edge structures of the scene as the maxima of the light density to produce a semi-dense depth map. Multiple viewpoints are then used to fuse the depth maps and complete 3D reconstruction. Guan et al. introduced a new hybrid tracking and dense mapping system based on neuromorphic cameras called EVI-SAM \cite{[T63]}. The pipeline of EVI-SAM includes two parallel modules: an Event Visual-Inertial Odometry (EVIO) module for tracking and estimating camera poses and an event-driven mapping module.

\subsection{Deep Learning-based Methods}
Recent methods using monocular neuromorphic cameras for 3D reconstruction rely on existing synthetic or real event datasets to train deep learning models for dense 3D reconstruction results. However, pipelines and the estimation of physical priors remain indispensable \cite{[survey]}. 

Baudron et al. proposed the E3D method, including a pipeline with a neural network \cite{[T10]}, E2S, to convert event frames into contours while using an additional neural branch for camera pose regression, ultimately generating multi-view mesh reconstruction results. Xiao et al. utilized the E2VID \cite{[T13-12]} deep learning method to process continuous event streams and output normalized intensity image sequences \cite{[T13]}. They used Structure-from-Motion (SfM) to estimate sparse intrinsic, extrinsic point clouds, followed by Multi-View Stereo (MVS) techniques to complete dense reconstruction. Wang et al. proposed a method that includes a physical prior extraction branch and a NeRF \cite{[T64]} rendering branch \cite{[T14]}. By incorporating an event warping field and a deterministic event generation model, they integrated physical priors into the NeRF pipeline. They also introduced a novel probabilistic chunk sampling strategy based on spatial event density, which helps the model learn local geometric features more robustly and efficiently.

The study done by Chen et al. in 2023 serves as the baseline for this research \cite{[Dense]}. It is the first study to use the voxel grid to represent event-based 3D reconstruction results. The study primarily introduces a deep learning model, E2V, which consists of a 3D event frame encoder to transform the feature representation of the data and a 3D voxel decoder to convert these features into a 3D voxel grid. Additionally, this research proposed a dataset, SynthEVox3D, for event-based 3D reconstruction. This method opened up a new direction for event-based 3D reconstruction, while its reconstruction accuracy on mIoU was only 0.346, which shows room for further improvement.
\begin{figure*}[ht] 
\centering
\includegraphics[width=\textwidth]{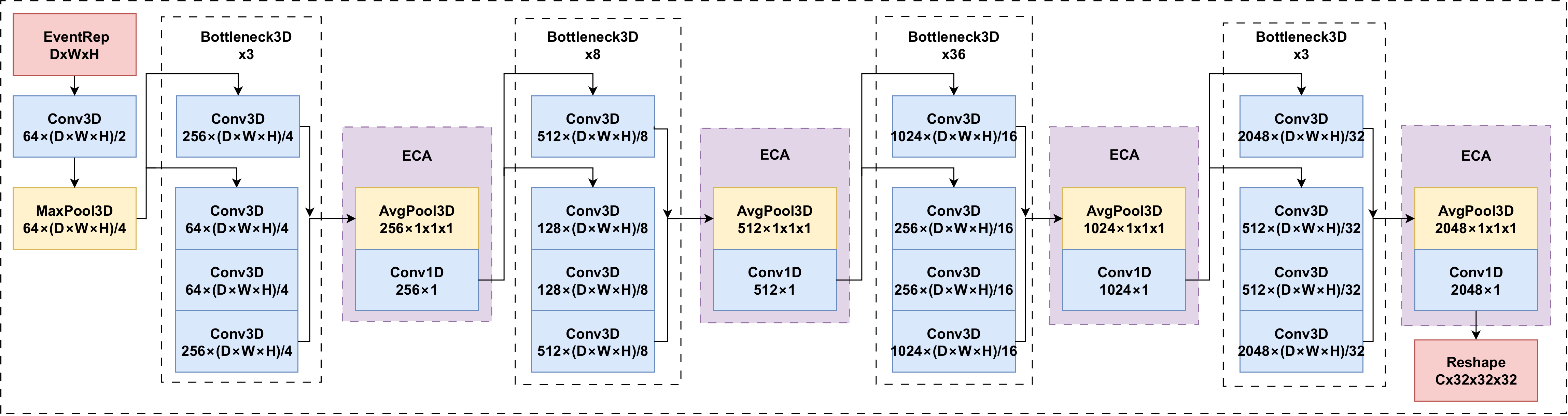}
\caption{Enhanced 3D ResNet Encoder.}
\label{fig: model}
\end{figure*}

\section{Method}
We propose an end-to-end method for dense voxel 3D reconstruction using monocular neuromorphic camera that eliminates the need to estimate physical priors.
\subsection{Novel Event Representation: Sobel Event Frame}

Event representation refers to preprocessing the event data, which encodes the event’s coordinate, timestamp, and polarity, aiming for better feature extraction \cite{[eventrep1], [eventrep2]}. 

Event Frame is the most commonly used event representation method \cite{[R3]}, referring to all approaches that can frame event streams. Event Frame divides event data into multiple fixed-length time windows, either based on timestamp or event count. Each time window generates an image-like frame where each pixel represents whether an event occurred within that time window. Pixels where no event occurred during the time window are filled with 0.

Each event frame \( F_t(x, y) \) is a pixel value generated in the time window \( t \), where \( (x, y) \) represents the pixel coordinates. \( E_i(x, y) \) represent the \( i \)-th event in the event stream within time window \( t \). \( p_i \) denote the polarity of event \( E_i \) (with \( +1 \) for positive polarity and \( -1 \) for negative polarity). \( n \) is the total number of events in the time window \( t \).

We summarize all five design modes of Event Frame as follows: 
\paragraph{Last Positive Event (Pos)} If the last event at each pixel within the time window has a positive polarity, it is marked as 1, otherwise 0.
\[
F_t(x, y) =
\begin{cases} 
1, & \text{if } p_n = +1, \\
0, & \text{if } p_n \neq +1.
\end{cases}
\]
\paragraph{Last Negative Event (Neg)} If the last event at each pixel within the time window has a negative polarity, it is marked as 1, otherwise 0.
\[
F_t(x, y) =
\begin{cases} 
1, & \text{if } p_n = -1, \\
0, & \text{if } p_n \neq -1.
\end{cases}
\]
\paragraph{Last Event Polarity (Last)} If the last event at each pixel within the time window has a positive polarity, it is marked as 1, and if negative, it is marked as -1.
\[
F_t(x, y) = p_n.
\]
\paragraph{Any Event (Any)} If any event occurs at a pixel within the time window, regardless of polarity, it is marked as 1.
\[
F_t(x, y) =
\begin{cases} 
1, & \text{if } n > 0, \\
0, & \text{if } n = 0.
\end{cases}
\]
\paragraph{Separate Frames for Polarity (Sep)} For each time window, a positive event frame $F_t^+(x, y)$ is generated for positive events in a time window, and a negative event frame $F_t^-(x, y)$ is generated for negative events in a time window.
\[
F_t^+(x, y) = 
\begin{cases} 
1, & \text{if any } p_i = +1, \\
0, & \text{otherwise}.
\end{cases}
\]
\[
F_t^-(x, y) = 
\begin{cases} 
1, & \text{if any } p_i = -1, \\
0, & \text{otherwise}.
\end{cases}
\]

The Sobel operator is an image-processing algorithm primarily used for edge detection. It computes the gradient of pixel intensity in both the horizontal (x-axis) and vertical (y-axis) directions using convolution, aiming to highlight the edges in the image \cite{[T74]}.

In neuromorphic camera-related research, there has been some work on applying convolution to event data, but only a few studies have used the Sobel operator \cite{[T76],[T77]}, and these are applied on initial event stream aiming to detect edge and structure in dynamic scenes, which can be considered as a non-frame-based preprocessing method for events.

The Sobel operator has never been applied to frame-based event representations in the neuromorphic camera field. Our attempt introduces the Sobel operator to continuously highlight edges of patterns in the event frame for 3D reconstruction, and we name this method \textit{Sobel Event Frame}. Applying \textit{Sobel Event Frame} to all pixels continuously can be represented by the following formula:
\begin{equation}
    O(x, y, t) = \sum_{w=-a}^{a} \sum_{h=-b}^{b} E(x + w, y + h, t) \cdot S(w, h)
\end{equation}

Here, \( E(x, y, t) \) denotes the event frame at time window \( t \). \( O(x, y, t) \) is the output frame after applying the convolution. \( S(w, h) \) is the convolution kernel. \( a \) and \( b \) are half the height $h$ and width $w$ of the convolution kernel, respectively. We select a kernel size $m=n=3$ in experiments. After extracting the gradient magnitude, we normalize it to the greyscale intensity range $[0,255]$ for visualization. Referring to Figure \ref{fig:sobel}, we can intuitively perceive the pattern's edges are enhanced.

\begin{figure*}[!h]
    
    \foreach \i in {1,3,4,5,6,7,8,9,10,11,12,13} {
        \begin{subfigure}{0.333\linewidth}
            \centering
            \includegraphics[width=\linewidth]{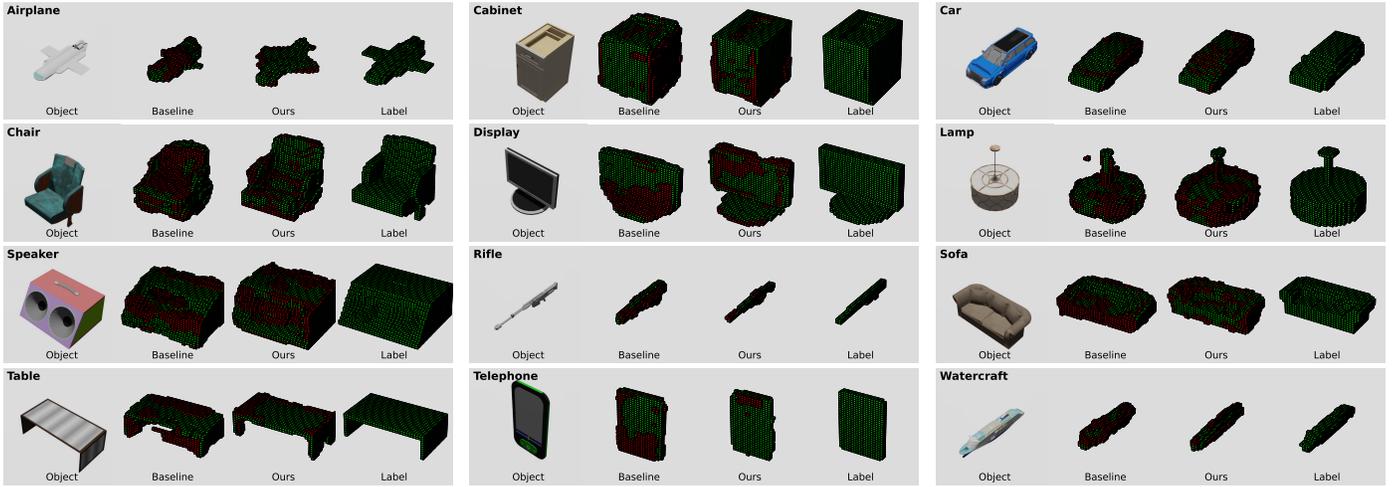}
            
        \end{subfigure}
        \ifnum\i=4\relax\\\fi
        \ifnum\i=7\relax\\\fi
        \ifnum\i=10\relax\\\fi
    }
    \caption{Visualization of 3D voxel reconstruction results.}
    \label{fig:result_voxel}
\end{figure*}

\subsection{Event-to-Voxel 3D ResNet with ECA}
We propose a model to process event data to dense voxel 3D reconstruction, effectively incorporating the event data represented by \textit{Sobel Event Frame} and further enhancing the learning on edge feature information.

The E2V model uses a deeper model architecture, ResNet-152, as the encoder to extract complex 3D features from the represented event stream \cite{[Dense]}. However, due to massive information in 3D data, the extracted features may include irrelevant or redundant information (e.g., the object is primarily located in the center of the event frame, while no events occur around the edges), which limits the performance of the model.

Inspired by Wang et al. \cite{[Chen1]}, we introduce the Efficient Channel Attention (ECA) mechanism into the encoder with deeper convolutional layers. This mechanism aims to efficiently capture inter-channel interactions by employing a 1D convolution with an adaptively determined kernel size, thus avoiding the dimensional reduction seen in other channel attention mechanisms. It achieves high performance without increasing the complexity of the deep CNN model.

Referring to Figure \ref{fig: model}, we integrate an ECA module after each bottleneck layer of the ResNet-152 encoder, ensuring that the model can focus on more meaningful information at every stage of feature extraction. We expect edge features strengthened by the \textit{Sobel Event Frame} will be enhanced again. The ECA module first applies global average pooling to each channel, capturing the global context of the input 3D feature maps. Then, a lightweight 1D convolution is applied to the pooled features to learn the dependencies among channels. A sigmoid activation function follows, normalizing the attention scores to values between 0 and 1. This process is highly efficient because it avoids fully connected layers and directly applies attention through a 1D convolution.

For the decoder, we add a controllable output reshaping module to E2V decoder \cite{[Dense]}, learning the correlation between the event data and ground truth labels, to generate the voxel logit output.

\begin{table*}[ht]
\centering
\rowcolors{2}{blue!3}{blue!10}
\caption{IoU results for each category}
\resizebox{\textwidth}{!}{%
\begin{tabular}{c|c|c|ccccccccccccc}
\hline
\textbf{Event Representation} & \textbf{Threshold} & \textbf{   mIoU   } & \textbf{Rifle} & \textbf{Chair} & \textbf{Car} & \textbf{Table} & \textbf{Sofa} & \textbf{Spkr} & \textbf{Airpln} & \textbf{Dspl} & \textbf{Wtrcft} & \textbf{Lamp} & \textbf{Cbn} & \textbf{Bench} & \textbf{Tel} \\
\hline
E2V - Event Frame (Pos) & 0.22 & 0.358 & 0.384 & 0.318 & 0.446 & 0.346 & 0.403 & 0.327 & 0.442 & 0.262 & 0.398 & 0.311 & 0.417 & 0.352 & 0.254 \\

E2V - Event Frame (Any) & 0.19 & 0.377 & 0.393 & 0.358 & 0.478 & 0.284 & 0.411 & 0.320 & 0.453 & 0.306 & 0.445 & 0.326 & 0.406 & 0.324 & 0.393 \\

Ours - Sobel EvtFrm (Pos) & 0.34 & \textcolor{red}{\textbf{0.523}} & 0.610 & 0.451 & 0.598 & 0.443 & 0.523 & 0.461 & 0.514 & 0.535 & 0.524 & 0.440 & 0.617 & 0.514 & 0.569 \\

Ours - Sobel EvtFrm (Any) & 0.33 & 0.512 & 0.655 & 0.437 & 0.569 & 0.451 & 0.539 & 0.458 & 0.570 & 0.607 & 0.515 & 0.372 & 0.548 & 0.372 & 0.560 \\
\hline
\end{tabular}%
}
\label{table: IoU results for each category}
\end{table*}

\begin{table*}[th]
\centering
\rowcolors{2}{blue!3}{blue!10}
\caption{F-Score results for each category}
\resizebox{\textwidth}{!}{%
\begin{tabular}{c|c|c|ccccccccccccc}
\hline
\textbf{Event Representation} & \textbf{Threshold} & \textbf{F-Score} & \textbf{Rifle} & \textbf{Chair} & \textbf{Car} & \textbf{Table} & \textbf{Sofa} & \textbf{Spkr} & \textbf{Airpln} & \textbf{Dspl} & \textbf{Wtrcft} & \textbf{Lamp} & \textbf{Cbn} & \textbf{Bench} & \textbf{Tel} \\
\hline
E2V - Event Frame (Pos) & 0.22 & 0.507 & 0.544 & 0.465 & 0.616 & 0.494 & 0.559 & 0.467 & 0.605 & 0.382 & 0.557 & 0.451 & 0.576 & 0.502 & 0.380 \\
E2V - Event Frame (Any) & 0.19 & 0.532 & 0.552 & 0.516 & 0.640 & 0.421 & 0.578 & 0.471 & 0.620 & 0.447 & 0.602 & 0.478 & 0.565 & 0.476 & 0.555 \\
Ours - Sobel EvtFrm (Pos) & 0.34 & \textcolor{red}{\textbf{0.682}} & 0.767 & 0.614 & 0.755 & 0.593 & 0.689 & 0.625 & 0.681 & 0.695 & 0.679 & 0.604 & 0.769 & 0.679 & 0.723 \\
Ours - Sobel EvtFrm (Any) & 0.33 & 0.667 & 0.796 & 0.596 & 0.726 & 0.616 & 0.702 & 0.610 & 0.728 & 0.757 & 0.675 & 0.528 & 0.695 & 0.520 & 0.719 \\
\hline
\end{tabular}%
}
\label{table: F-Score results for each category}
\end{table*}

\subsection{Optimal Binarization Threshold Selection Principle}

After generating the model, we use it to perform inference (prediction) on the testing data subset. The output of the model consists of continuous logits, which will be converted into continuous probability values $\sigma(x)$ by Sigmoid function. By setting a binarization threshold $p$, we can determine whether a reconstructed point is present or absent as follows:

\begin{equation}
\text{Output Binary Value} = 
\begin{cases} 
1, & \text{if } \sigma(x_{\text{out}}) > p, \\
0, & \text{if } \sigma(x_{\text{out}}) \leq p.
\end{cases}
\end{equation}
\begin{equation}
\text{,where} \quad \sigma(x_{\text{out}}) = \frac{1}{1 + e^{-x_{\text{out}}}}
\end{equation}

Previous studies used a fixed binarization threshold such as 20\% \cite{[Dense]} or 50\% \cite{[T10]}. However, we found that the optimal threshold varies depending on the event representation method. 

To determine the best threshold $p^*$, we iterate through a predefined range of thresholds $P$ (from 0.15 to 0.50 with a step size of 0.01) and select the threshold that yields the best mIoU and F-Score as follows:

\begin{equation}
    p^* = \underset{p \in P}{\arg\max} \, 
    \begin{cases} 
        \text{mIoU}(p), & \text{for mIoU} \\
        \text{F-Score}(p), & \text{for F-Score}
    \end{cases}
\end{equation}

\section{Experiments}
\subsection{Datasets and Implementation Details}
\textbf{Dataset}: Synthetic Event Camera Voxel 3D Reconstruction Dataset (SynthEVox3D) is the only available dataset for 3D reconstruction with voxel label based on event data. SynthEVox3D consists of 39,739 event data samples in 13 categories sourced from ShapeNet \cite{[T71]}, including Airplane, Bench, Cabinet, Car, Chair, Displayer, Lamp, Speaker, Rifle, Sofa, Table, Telephone, and Watercraft. Each category contains objects of different shapes, ensuring data differentiation. The event data are generated by all-angle scanning within $0.5s$ using a neuromorphic camera, producing event streams with a resolution of $512 \times 512$ pixels. SynthEVox3D-Tiny is a subset of SynthEVox3D, which randomly selects 80 samples from each category, making a total of 1,040 samples. These samples include 832 training data (80\%), 104 validation data (10\%), and 104 testing data (10\%).

\textbf{Event Representation}: We use a fixed time window of $5.0 \times 10^{-3}$ to segment the event data into event frames, forming 100 frames (total duration of 0.5s). The karnel size of \textit{Sobel Event Frame} is set to $3 \times 3$. After event representation, the event data is normalized to $[0, 1]$ and rescaled to a shape of $(n, 256, 256)$ for fitting the model input. To further prevent overfitting, we randomly apply operations such as flipping, rotation, and inversion of the event stream.

\textbf{Deep Learning}: We set the batch size to $5$ and use the Adam optimizer with a learning rate of $1.0 \times 10^{-6}$. The optimizer is adopted with $\beta = \{0.9, 0.999\}$. The total number of training epochs is set to 100. The dropout rate is set to $0.25$. The loss function used is Focal Loss due to the imbalanced data structure (polarization data).

\subsection{Experimental Processes \& Results}
First, we conducted training on the SynthEVox3D-Tiny dataset. We reproduced the E2V method using \textit{Event Frame (Pos)} and evaluated it on the testing data subset as the baseline. At the optimal threshold of 0.22, the E2V method achieved mIoU of 0.358 and F-Score of 0.507. 

Subsequently, we tested E2V on the other four modes of \textit{Event Frame} and evaluated all five modes of \textit{Sobel Event Frame} using our method. The \textit{(Pos)} and \textit{(Any)} modes, which do not contain negative data, performed better under the current experimental settings. Therefore, in Tables \ref{table: IoU results for each category} and \ref{table: F-Score results for each category}, we only present the results of them.

Using \textit{Sobel Event Frame (Pos)} with a binarization threshold of 0.34, our method achieved mIoU of 0.523 and F-Score of 0.682. This represents an improvement of 0.165 (46.1\%) in mIoU and 0.175 (34.5\%) in F-Score compared to the E2V method with \textit{Event Frame (Pos)}.

\begin{table}[h]
\caption{Comparison of mIoU scores across methods.}
\centering
\begin{tabular}{lcc}
\hline
                        & \textbf{Training Data Size} & \textbf{mIoU} \\
\hline
3D-R2N2                 &                            & 0.636         \\
AttSets                 &  \multirow{2}{*}{$\sim$30,000}               & 0.693         \\
Pix2Vox++               &                            & 0.706         \\
EVolT                   &                            & 0.735         \\
\hline
E2V (Tiny)             & 832                        & 0.358         \\
E2V (Full)             & 39,739                        & 0.346         \\
Ours (Tiny)               & 832                       & 0.523       \\
\rowcolor{green!10}
Ours (Full)               & 39,739                       & \textcolor{red}{\textbf{0.535}}       \\
\hline
\end{tabular}
\label{table:comparison_miou}
\end{table}
Then, we conducted a full dataset evaluation on SynthEVox3D with our model and the best-performing mode: \textit{Sobel Event Frame (Pos)}. As shown in Table \ref{table:comparison_miou}, our method achieved a mIoU of 0.535 on the full dataset, which represents a 54.6\% improvement over E2V, demonstrating the stability of our method even with extensive training data. We also compared our method with state-of-the-art voxel reconstruction methods based on multi-view traditional images. The mIoU values of these traditional methods were obtained from 20 views of objects in the ShapeNet dataset. While the datasets are not entirely comparable, our reconstruction accuracy is approaching that of these traditional methods. We believe that if both methods were compared under the same extremely rapid scanning conditions, traditional methods will suffer from motion blur, but our method will perform better, which we plan to test as a future work.

Figure \ref{fig:result_voxel} presents visualizations of 12 random reconstruction results from the testing data, showing each object, voxel reconstruction results from our method and baseline, and ground truth labels. In the voxel reconstruction results, we use green to represent correctly reconstructed voxels and red to represent incorrectly reconstructed voxels. It can be observed that our voxel outputs perform better than the baseline. There are no outlier voxels from incorrect reconstructions surrounding the object. The category of the object can be easily identified.

\subsection{Ablation Study}
Our method can be divided into three components: event representation, model, and binarization threshold selection, for conducting ablation studies to evaluate the contribution of each component to the reconstruction results.

\begin{table}[h]
    \centering
    \caption{Ablation Study Results on Event Representation and Model Variants}
    \begin{tabular}{cccc}
    \hline
        \textbf{EventRep} & \textbf{Model} & \textbf{Threshold} & \textbf{mIoU} \\
        \hline
        Event Frame (Pos) & E2V & 0.22 & 0.358 \\
        Event Frame (Pos) & Ours & 0.21 & 0.421 (17.6\%↑)\\
        Sobel EvtFrm (Pos) & E2V & 0.33 & 0.471 (31.6\%↑) \\
        \rowcolor{green!10}
        Sobel EvtFrm (Pos) & Ours & 0.34 & \textcolor{red}{\textbf{0.523 (46.1\%↑)}} \\
        \hline
    \end{tabular}
    \label{tab:ablation_study}
\end{table}

\textbf{Event Representation and Model:} We conducted ablation studies on event representation and the model using the SynthEVox3D-Tiny dataset, referring to Table \ref{tab:ablation_study} for the results. By using \textit{Event Frame (Pos)} to process event data as input to our proposed model, we obtained a mIoU of 0.421 at a threshold of 0.21, which means our model brought about a 17.6\% improvement. Using the E2V model, we also trained and tested with \textit{Sobel Event Frame (Pos)}, achieving a mIoU of 0.471 at a threshold of 0.33. In other words, relying solely on \textit{Sobel Event Frame} resulted in a 31.6\% improvement. The combination of both led to a further enhancement, reaching a mIoU of 0.523, a 46.1\% improvement over the baseline.

\begin{table}[t]
\centering
\caption{The mIoU and F-Score results of \textit{Sobel Event Frame (Pos)} under different binarization thresholds.}
\begin{tabular}{ccc|ccc}
\hline
\textbf{Threshold} & \textbf{mIoU} & \textbf{F-Score} & \textbf{Threshold} & \textbf{mIoU} & \textbf{F-Score} \\
\hline
0.16 & 0.488 & 0.653 & 0.30 & 0.521 & 0.681 \\

0.18 & 0.496 & 0.660 & 0.32 & 0.522 & 0.682  \\

0.20 & 0.503 & 0.666 & \cellcolor{green!10}\textcolor{red}{\textbf{0.34}} & \cellcolor{green!10}\textcolor{red}{\textbf{0.523}} & \cellcolor{green!10}\textcolor{red}{\textbf{0.682}} \\

0.22 & 0.508 & 0.670 & 0.36 & 0.522 & 0.681 \\

0.24 & 0.513 & 0.674 & 0.38 & 0.521 & 0.680 \\

0.26 & 0.516 & 0.677 & 0.40 & 0.519 & 0.677  \\

0.28 & 0.519 & 0.679 & 0.42 & 0.516 & 0.674  \\

\hline
\end{tabular}%
\label{tab:Different thresholds}
\end{table}

\textbf{Binarization Threshold:} To demonstrate that the binarization threshold should be optimally selected rather than fixed, we listed the mIoU and F-Score results using \textit{Sobel Event Frame (Pos)} at thresholds ranging from 0.16 to 0.42 (with a step size of 0.02), as shown in Table \ref{tab:Different thresholds}. The best results were achieved at a threshold of 0.34, reaching a mIoU of 0.523 and an F-Score of 0.682. Performance was slightly worse at thresholds of 0.32 and 0.36. Deviating from the threshold of 0.34 in either direction led to poorer results. The E2V method suggests using 0.20 as a reference threshold \cite{[Dense]}, but this did not yield the best results in our experiments. Within this range, using the worst threshold (0.16) would result in a mIoU decrease of 0.035. Additionally, according to Table \ref{tab:ablation_study}, applying different event representation methods had a significant impact on the optimal threshold, while the model only had a smaller effect on the optimal threshold.

Therefore, this demonstrates the significant impact of the binarization threshold on model performance, indicating that the optimal threshold is not fixed but needs to be optimized based on the specific event representation method. We recommend that future related tasks refer to our \textit{Optimal Binarization Threshold Selection Principle}, flexibly adjusting the threshold to optimize results, which serves as a guideline for future work.

\section{Conclusion \& Future Work}
In this study, we proposed a novel event representation method, \textit{Sobel Event Frame}, and an ECA-enhanced deep learning method for end-to-end 3D reconstruction using the monocular neuromorphic camera without relying on physical priors and pipelines. Additionally, we introduced a \textit{Optimal Binarization Threshold Selection Principle} and advocated for it as a guideline for future event-based 3D reconstruction. Under our approach, we achieved a significant improvement in reconstruction accuracy, further bridging the gap with results from traditional camera-based 3D reconstruction.

Looking ahead, we plan to expand this project further. Regarding event representation, our aim is to explore more potential representations and extend \textit{Sobel Event Frame} to other computer vision tasks. Although we introduced \textit{Optimal Binarization Threshold Selection Principle}, it is still necessary to investigate the factors that influence threshold selection. Most importantly, we need to test the performance of traditional methods and event-based methods under extreme conditions, such as very fast motion, low light, or high brightness scenarios, which will demonstrate the advantages of neuromorphic cameras in the field of 3D reconstruction.

\bibliographystyle{IEEEbib}
\bibliography{output}

\begin{thebibliography}{10}

\bibitem{[T7]}
G.~Gallego, T.~Delbrück, G.~Orchard, C.~Bartolozzi, B.~Taba, A.~Censi, S.~Leutenegger, A.~J. Davison, J.~Conradt, K.~Daniilidis, et~al.,
\newblock ``Event-based vision: A survey,''
\newblock {\em IEEE Transactions on Pattern Analysis and Machine Intelligence}, vol. 44, no. 1, pp. 154--180, 2020.

\bibitem{[T13-12]}
H.~Rebecq, R.~Ranftl, V.~Koltun, and D.~Scaramuzza,
\newblock ``High speed and high dynamic range video with an event camera,''
\newblock {\em IEEE Transactions on Pattern Analysis and Machine Intelligence}, vol. 43, no. 6, pp. 1964--1980, 2019.

\bibitem{qu2024e2hqv}
Q.~Qu, Y.~Shen, X.~Chen, Y.~Y. Chung, and T.~Liu,
\newblock ``E2hqv: High-quality video generation from event camera via theory-inspired model-aided deep learning,''
\newblock in {\em AAAI}, 2024, vol.~38, pp. 4632--4640.

\bibitem{[survey]}
C.~Xu, H.~Zhou, L.~Chen, H.~Chen, Y.~Zhou, V.~Chung, Q.~Qu, and W.~Cai,
\newblock ``A survey of 3d reconstruction with event cameras,''
\newblock {\em arXiv preprint arXiv:2505.08438}, 2025.

\bibitem{[T5]}
H.~Rebecq, G.~Gallego, E.~Mueggler, and D.~Scaramuzza,
\newblock ``Emvs: Event-based multi-view stereo—3d reconstruction with an event camera in real-time,''
\newblock {\em International Journal of Computer Vision}, vol. 126, no. 12, pp. 1394--1414, 2018.

\bibitem{[T10]}
A.~Baudron, Z.~W. Wang, O.~Cossairt, and A.~K. Katsaggelos,
\newblock ``E3d: event-based 3d shape reconstruction,''
\newblock {\em arXiv:2012.05214}, 2020.

\bibitem{[T13]}
K.~Xiao, G.~Wang, Y.~Chen, J.~Nan, and Y.~Xie,
\newblock ``Event-based dense reconstruction pipeline,''
\newblock in {\em 2022 ICRAS}. IEEE, 2022, pp. 172--177.

\bibitem{[T14]}
J.~Wang, J.~He, Z.~Zhang, and R.~Xu,
\newblock ``Physical priors augmented event-based 3d reconstruction,''
\newblock pp. 16810--16817, 2024.

\bibitem{[T9]}
H.~Rebecq, T.~Horstschaefer, G.~Gallego, and D.~Scaramuzza,
\newblock ``Evo: A geometric approach to event-based 6-dof parallel tracking and mapping in real time,''
\newblock {\em IEEE Robotics and Automation Letters}, vol. 2, no. 2, pp. 593--600, 2016.

\bibitem{[Dense]}
H.~Chen, V.~Chung, L.~Tan, and X.~Chen,
\newblock ``Dense voxel 3d reconstruction using a monocular event camera,''
\newblock in {\em 2023 ICVR}. IEEE, 2023, pp. 30--35.

\bibitem{[T6]}
H.~Kim, S.~Leutenegger, and A.~J. Davison,
\newblock ``Real-time 3d reconstruction and 6-dof tracking with an event camera,''
\newblock in {\em ECCV}. Springer, 2016, pp. 349--364.

\bibitem{[T61]}
J~Conradt,
\newblock ``On-board real-time optic-flow for miniature event-based vision sensors,''
\newblock in {\em 2015 IEEE International Conference on Robotics and Biomimetics (ROBIO)}. IEEE, 2015, pp. 1858--1863.

\bibitem{[T63]}
W.~Guan, P.~Chen, H.~Zhao, Y.~Wang, and P.~Lu,
\newblock ``Evi-sam: Robust, real-time, tightly-coupled event--visual--inertial state estimation and 3d dense mapping,''
\newblock {\em Advanced Intelligent Systems}, vol. 6, no. 12, pp. 2400243, 2024.

\bibitem{[T64]}
B.~Mildenhall, P.~P. Srinivasan, M.~Tancik, J.~T. Barron, R.~Ramamoorthi, and R.~Ng,
\newblock ``Nerf: Representing scenes as neural radiance fields for view synthesis,''
\newblock {\em Communications of the ACM}, vol. 65, no. 1, pp. 99--106, 2021.

\bibitem{[eventrep1]}
Q.~Qu, H.~Liang, X.~Chen, Y.~Y. Chung, and Y.~Shen,
\newblock ``Nerf-nqa: No-reference quality assessment for scenes generated by nerf and neural view synthesis methods,''
\newblock {\em IEEE TVCG}, 2024.

\bibitem{[eventrep2]}
Z.~Yu, Q.~Qu, Q.~Zhang, N.~Zhang, and X.~Chen,
\newblock ``Llm-evrep: Learning an llm-compatible event representation using a self-supervised framework,''
\newblock {\em arXiv preprint arXiv:2502.14273}, 2025.

\bibitem{[R3]}
X.~Zheng, Y.~Liu, Y.~Lu, T.~Hua, T.~Pan, W.~Zhang, D.~Tao, and L.~Wang,
\newblock ``Deep learning for event-based vision: A comprehensive survey and benchmarks,''
\newblock {\em arXiv:2302.08890}, 2023.

\bibitem{[T74]}
N.~Kanopoulos, N.~Vasanthavada, and R.~L Baker,
\newblock ``Design of an image edge detection filter using the sobel operator,''
\newblock {\em IEEE Journal of solid-state circuits}, vol. 23, no. 2, pp. 358--367, 1988.

\bibitem{[T76]}
C.~Brändli, J.~Strubel, S.~Keller, D.~Scaramuzza, and T.~Delbruck,
\newblock ``Elised—an event-based line segment detector,''
\newblock in {\em 2016 EBCCSP}. IEEE, 2016, pp. 1--7.

\bibitem{[T77]}
C.~Scheerlinck, N.~Barnes, and R.~Mahony,
\newblock ``Asynchronous spatial image convolutions for event cameras,''
\newblock {\em IEEE Robotics and Automation Letters}, vol. 4, no. 2, pp. 816--822, 2019.

\bibitem{[Chen1]}
Q.~Wang, B.~Wu, P.~Zhu, P.~Li, W.~Zuo, and Q.~Hu,
\newblock ``Eca-net: Efficient channel attention for deep convolutional neural networks,''
\newblock in {\em CVPR}, 2020, pp. 11534--11542.

\bibitem{[T71]}
A.~X. Chang, T.~Funkhouser, L.~Guibas, P.~Hanrahan, Q.~Huang, Z.~Li, S.~Savarese, M.~Savva, S.~Song, H.~Su, and Others,
\newblock ``Shapenet: An information-rich 3d model repository,''
\newblock {\em arXiv preprint arXiv:1512.03012}, 2015.

\end{thebibliography}

\end{document}